\documentclass[10pt]{article}

\usepackage[final]{ap_2017}

\usepackage[utf8]{inputenc} % allow utf-8 input
\usepackage[T1]{fontenc}    % use 8-bit T1 fonts
\usepackage{hyperref}       % hyperlinks
\usepackage{url}            % simple URL typesetting
\usepackage{booktabs}       % professional-quality tables
\usepackage{amsfonts}       % blackboard math symbols
\usepackage{nicefrac}       % compact symbols for 1/2, etc.
\usepackage{microtype}      % microtypography
\usepackage{times}
\usepackage{graphicx}
\graphicspath{{images/}}
\usepackage{natbib}

\usepackage{datetime2}
\usepackage{wrapfig}
\usepackage{siunitx}
\usepackage{verbatim}
\usepackage{fancyhdr}
\title{RubyStar: A Non-Task-Oriented Mixture Model Dialog System}

\author{
	Huiting Liu, Tao Lin, Hanfei Sun, Weijian Lin, Chih-Wei Chang, Teng Zhong, Alexander Rudnicky\\
	School of Computer Science, Carnegie Mellon University\\
	5000 Forbes Avenue, Pittsburgh, PA 15213\\
	\texttt{\{huitingl, tao.lin, hanfeis, wlin1, cchang3, tzhong, air\}@cs.cmu.edu} \\     
}

\begin{document}

\maketitle

\begin{abstract}
RubyStar is a dialog system designed to create ``human-like'' conversation by combining different response generation strategies. RubyStar conducts a non-task-oriented conversation on general topics by using an ensemble of rule-based, retrieval-based and generative methods. Topic detection, engagement monitoring, and context tracking are used for managing interaction. Predictable elements of conversation, such as the bot's backstory and simple question answering are handled by separate modules. We describe a rating scheme we developed for evaluating response generation. We find that character-level RNN is an effective generation model for general responses, with proper parameter settings; however other kinds of conversation topics might benefit from using other models.
\end{abstract}

\section{Introduction}
The implicit goal of an open-domain non-task-oriented dialog system is to generate a coherent, pleasant and engaging conversation responsive to a user's inputs.  In retrieval-based approaches, large databases provide candidate responses. Recent advances in neural networks (\citet{mikolov2010recurrent}), especially sequence to sequence frameworks (\citet{sutskever2014sequence}), have lead to generative-based approaches. Both approaches have its own drawbacks: (i) retrieval-based methods depend highly  on the quality and scale of its underlying databases (ii) generative-based ones are more likely to give generic or inconsistent responses partly due to oversimplification of the objective function (\citet{vinyals2015neural}). An ensemble of these two approaches might inherit the advantages while overcoming the disadvantages of both (\citet{song2016two}).

In the real world,  user experience is critical to the success of non-task-oriented systems~(\citet{yu2016wizard}). Therefore, one should apply additional methods to guide the flow of conversation over successive contexts, including but not limited to: (i) using coreference resolution and sentiment analysis to achieve a more explicit representation of current context (ii) using a post-reranking model to select the coherent and engaging responses from candidates~(\citet{song2016two}) (iii) using backstories, handcrafted rules and templates to model reliable behavior for routine questions.

This paper describes RubyStar, a socialbot that conducts non-task-oriented conversations on general topics. RubyStar integrates traditional techniques to generate replies, such as template and information retrieval, modern techniques such as machine learning and deep learning models, as well as other available resources such as knowledge graph and QA engine. 
This architecture reflects our belief that human conversation is usefully conceived of as an assemblage of specific skills that are brought to bear as circumstance dictates. These might include ones that are task-specific, informative, social or simply chat. RubyStar represents an initial attempt to create a multi-faceted conversation capability.

The structure of this paper is as follows: we discuss related work in Section 2. Section 3 provides a system overview. Section 4 and 5 describes data collection and provides performance analysis. Follow-on work is described in Section 6.

\section{Related Work}

Interest in chat agents especially non-goal oriented socialbots have a long history, from early programs such as ELIZA (\citet{weizenbaum1966eliza}) and PARRY (\citet{colby1981}), to A.L.I.C.E.\footnote{\url{http://www.alicebot.org/}} and Microsoft's XiaoBing\footnote{\url{http://www.msxiaoice.com/}}, that have interacted with millions of users on social networking platforms. These chatbots can be categorized according to how responses are generated. Early chatbots were rule-based due to the simplicity of implementation, but hard-coded rules limit coverage. Corpus-based approaches to rule generation have been investigated (\citet{shawar2003using}) but limitations in generality were difficult to overcome. 

More recent corpus-based systems have tried to address the coverage issue. Retrieval-based approaches (\citet{banchs2012iris}) involve pre-indexed conversations and corpora. This approach has also been used for a Question Answering (QA) module in a dialog system (\citet {yan2016docchat}).

As deep learning approaches have achieved high performance in many NLP tasks (\citet{klein2017opennmt, hinton2012deep}), generative methods for chat have attracted interest (\citet{vinyals2015neural}). Recent work has mainly focused on improvement to context tracking in multi-turn conversation (\citet{sordoni2015neural, serban2015building}), diversity promotion to mitigate the \textit{Safe Response} issue (\citet{li2015diversity}), and persona consistency during response generation (\citet{li2016persona}).

Apart from these response generation paradigms, other work has explored strategies for keeping conversation engaging and extending an interaction for as long as possible. Such systems monitor user engagement (\citet{yu2015ticktock}) and can introduce new content or shift the topic proactively when they detect that users respond in an unmindful way (\citet{yu2016wizard,li2016stalematebreaker}).

Our own approach has focused on developing a socialbot that combines rule-based, retrieval-based and generative approaches, and on techniques that monitor a conversation in different ways with the goal of promoting continuity. These techniques include topic detection, context tracking, and engagement-based reranking.

It's worth mentioning that at least two other Alexa Prize teams' systems also use an ensemble of rule-based, retrieval-based and neural-based generative methods. \citet{BengioTeam} use a deep reinforcement learning approach to rerank candidate responses based on Amazon Mechanical Turk (AWT) crowd-sourcing data and online system's user interactions, while we use SVM for re-ranking based on Reddit up-voting data. \citet{2017arXiv170909816K} recruited AWT workers to write about 20,000 self-dialogues and use it as  data for their retrieval-based method, but also use rule-based and neural-based generative methods in the system. \citet{Slugbot} do not describe  their architecture, but discuss approaches to model retrieval-based discourse coherence and the potential to use reinforcement learning to improve the ranking of utterances.

\section{System Overview}
\begin{wrapfigure}{R}{10cm}
% \begin{figure*}
  \centering
  \includegraphics[width=10cm,height=6cm]{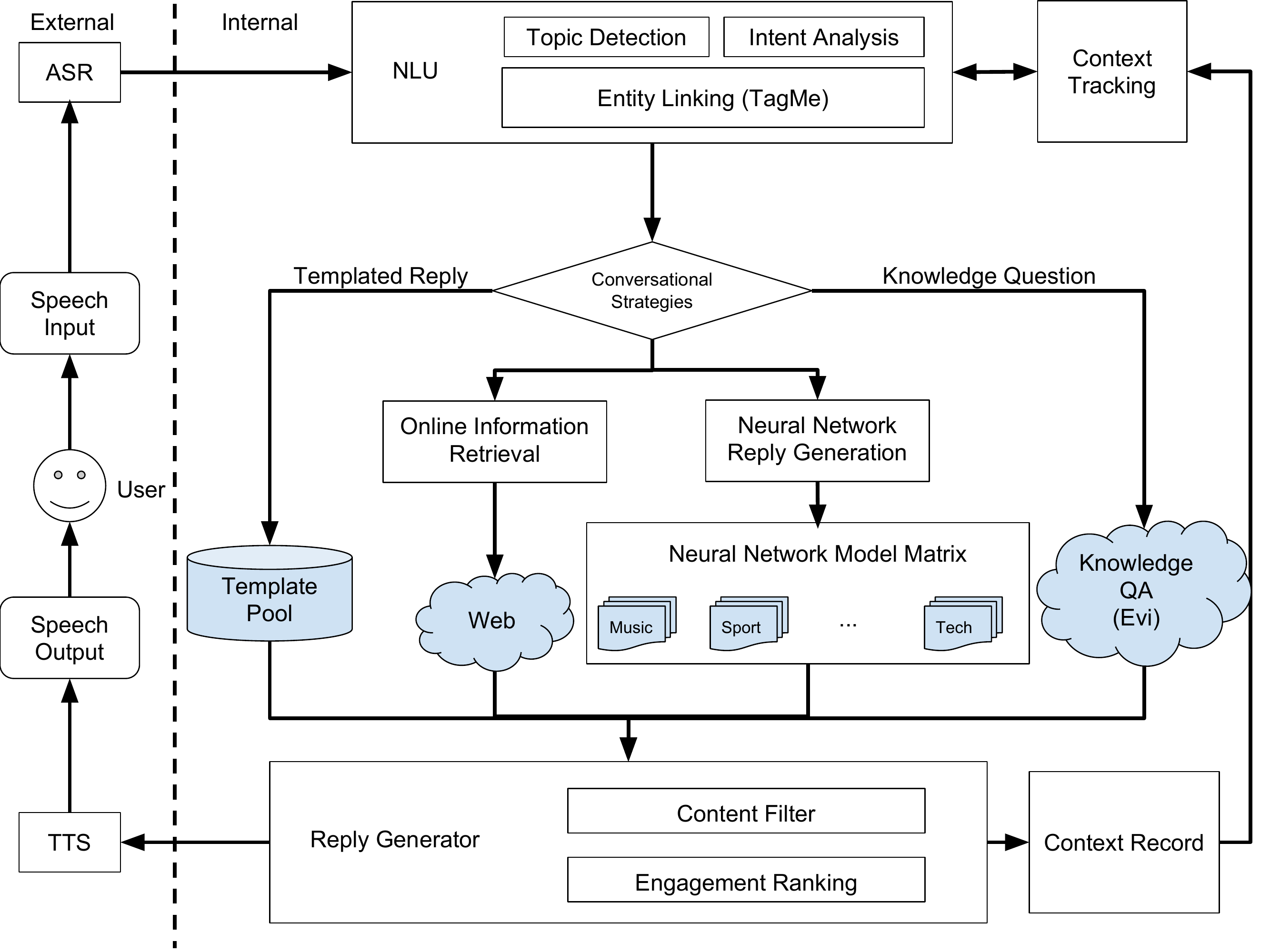}
  \caption{The architecture of RubyStar.}
\label{figure1}
\vspace{-10pt}
% \end{figure*}
\end{wrapfigure}

%\alex{I don't completely understand the following description. Can we talk about it?
%Also, the figure should put the Alexa bits on the vertical lines to the user, gives you a bit more room for your stuff (which in case should be larger so that %people can read stuff.)
%} 
Figure \ref{figure1} shows the architecture of the RubyStar Socialbot. The system uses Amazon Alexa ASR to decode speech input to text and Alexa TTS to generate spoken responses \footnote{\url{https://developer.amazon.com/alexa-voice-service}}.  

The Natural Language Understanding Unit (NLU) does preprocessing. It includes components for topic detection, intent analysis, as well as an entity linking.  The topic detection component computes a probability for each covered topic based on the sequence of preceding turns in the current conversation.  Intent analysis component will identify the user intent, so the system can handle the conversation with different strategies based on the intents. The entity linking component links entities in the input to entries in Wikipedia. Based on the NLU result, the system then will follow different paths, according to its built-in conversational strategies.

A strategies layer follows the NLU. We divide the strategies into four types. In  order of priority, the strategies are rule-based, knowledge-based, retrieval-based, and generative. The rule-based strategies are intent templates, backstory templates, and entity-based templates ordered by their priorities. Because rule-based strategies encode human knowledge into the form of templates, they provide the most accurate replies. The system will adopt a template reply if input is recognized by one of these strategies. If there is no matching template for the input, the system will try to get an answer from Evi, a knowledge-base QA engine provided by Amazon. Failing that, the input is handled by an ensemble of neural network models and information retrieval modules to create general conversation output. 

After going through one or more of strategies, the stream flows into the reply generator. The reply generator will first apply a content filter to eliminate incoherent or questionable candidates. If there are more than one valid replies, a ranking process is applied to select the best reply first according the priority and then the engagement ranking. Finally, the system will passes reply to the Text to Speech (TTS) to  generate the final output. Simultaneously, all conversations are tracked in a history. The context tracking module tries to improve coherence using the context and topic history of the current conversation to carry out coreference resolution and do topic weighting. 

Since the NLU and the reply generator don't change frequently, most of our work focused on the strategies layer. We start from simple rule-based strategies and then add more complicated strategies, such as knowledge-based strategies, retrieval-based strategies and generative strategies. We can also plug in or pull out one specific strategy to evaluate its contribution to system performance. 

In the remainder of this section, we  provide a more detailed description of each system component. 

\subsection{Intent Analysis}
To identify intents in conversation with the goal of selecting  appropriate replies, we classify each input into one of a set of intent categories.
We defined 42 common intents by labeling 10,000 user inputs from logs of users' interactions. Identified intents include some common backstory topics like favorite food, some common concepts such as yes or no, and  more complex ones such as instructions and opinion. As part of our workflow of defining intents and labeling inputs, we compiled an intent dictionary containing intent id, example, description, reply strategies. We then merged or split the dictionary intents to achieve an appropriate granularity of intent.

We then use Amazon Lex
~\footnote{\url{https://aws.amazon.com/lex/}}
\citet{DBLP:journals/corr/abs-1711-00549}
to map the input into an intent category. Examples for each intent were based on our corpus labeling results. Finally, we select different conversation strategies depending on the intent: for some intents, we  select randomly from prewritten templates; for others, we retrieve related discussion from Twitter.

\subsection{Topic Detection}
To guide the conversation on a more comfortable and engaging course, topic detection is used to track context and topics over time. The topic detector uses a random forest to classify the input sentence into one of several general topics. We selected Politics, Life, Sports, Entertainment, Technology and General; probabilities are generated for each topic. When a sentence is passed in, the module tokenizes the text and extracts informative keywords. The keywords are converted into word vectors and used as classifier features. While predicting the current topic, the classifier also takes previously detected topics into consideration. To minimize response time, we use a twenty tree random forest, and average the results over the last five turns.

Our training data comes from Reddit comments. Reddit pages are organized into area of interests, subreddits, and for the topic-based multi-model approach, we decide on the topic of a comment according to its subreddit title. Our model is trained on over 5,000 samples and tested on 500 samples, with an overall accuracy over 90\%.

\subsection{Entity-based Templates}
This module consists of an Named Entity Disambiguation (NED) model and a template selection model. NED (\cite{bunescu2006using,cucerzan2007large}) links entity mentioned in a text to a knowledge base, an essential step for allowing RubyStar to understand conversation topics and generate appropriate replies, as it connects words with concepts and subtexts in the real word using TAGME (\cite{ferragina2010tagme}). TAGME takes input text and return a list of entities with their Wikipedia titles, which in turn can be converted to nodes in the Wikidata (\cite{vrandevcic2012wikidata}) knowledge graph. Each entity includes a confidence score. An empirical threshold is used to select the top high-confidence entities. The cutoff of the confidence score is based on manually verifying the entities extracted from previous conversation logs.

After generating a list of entities, we use pre-authored templates to generate conversation replies using the template selection model. For each mentioned entity, we retrieve its attributes in Wikidata knowledge graph, to get its basic information as material for our reply. For example, if the entity is a movie actor, we retrieve an actor’s gender, age, acted films list; if the entity is a city, we retrieve its location, which country it belongs to, and famous people who lives in this city. Because of the limitation of the knowledge graph, sometimes not all related information is available. Based on information need for each template and the attributes we have for all the entities, we randomly select one from related templates to promote diversity in the conversation.

Here is an example of the process used to generate replies based on templates.
\begin{itemize}
\item User says ``I think Rush Hour is the best action movie I’ve ever seen''.
\item Use TAGME to find entities and link to Wikipedia, which is ``Rush Hour'' in this case.
\item Find the features of the entities by WikiData.
 Then query for the master entity ``Rush Hour'', find it has a feature type called ``cast member'', whose value is ``Jackie Chan''. This yields:
\begin{itemize}
\item \textbf{Master entity}: Rush Hour
\item \textbf{Feature type}: cast member
\item \textbf{Feature value}: Jackie Chan
\end{itemize}
\item Get the templates for the relation, fill in the pair of entities.
Given the templates for the relation ``<film, cast member>'' and the pair of entity ``<Rush Hour, Jackie Chan>'', it finds a template (which is manually collected).
\begin{itemize}
\item \textbf{Relation} <typeof(master entity), feature type> e.g. <film, cast member>
\item \textbf{Pair of entity} <master entity, feature value> e.g. <Rush Hour, Jackie Chan>
\item \textbf{Template} e.g. ``Last night I had a dream that I was [cast member]... So... I think I need to take a break from watching [film]''
\end{itemize}
So the system can fill in the blanks and replies ``Last night I had a dream that I was Jackie Chan... So... I think I need to take a break from watching Rush Hour''.
\end{itemize}

\subsection{Information Retrieval}
This module tries to provide more human-like, more concrete, and fresher replies compared to the entity-based template and neural dialog generation modules. Currently, the source of information for this module is the most recent tweets provided by Twitter search API~\footnote{\url{https://dev.twitter.com/overview/api}}. We use tweets as the source because they are usually short sentences closer to humans' verbal language compared to long written posts. Twitter data could also reflect trending topics and ideas quickly, compared to locally stored corpora. We are exploring additional information sources, such as Quora and Reddit, which however would require different selection strategies.

From the Twitter search API, the top one hundred (the number of tweets allowed by Twitter API) related English tweets in the recent seven days are retrieved. The keywords used for queries are based on the entities in the sentence, which are extracted by using TAGME (\cite{ferragina2010tagme}). Hashtags, emoticons, mentions, URLs and others are removed, as well as duplicate tweets. Considering that the language pattern on Twitter is sometimes different from English and unsuitable for a socialbot, the sentences with too many ``misspelled'' words are removed. The misspelled words might have special non-English characters, or have special patterns such as "It's sooooo hoooot!". Finally, a reply is randomly selected from the remained tweets. We are exploring learning  ranking methods to select more suitable replies.

For example, when a user asks ``\textit{How did Neil Gorsuch do in his confirmation hearings?}'', TagMe links ``\textit{Neil Gorsuch}'' to ``\textit{Neil Gorsuch}'' in Wikipedia, and ``\textit{confirmation}'' is linked to ``\textit{Advice and consent}''. So we send query ``\textit{("Neil Gorsuch" OR "Neil Gorsuch") ("Advice and consent" OR "confirmation")}'' to Twitter, and get the top 100 replies. After cleaning and removing duplicates, 48 replies remain. Replies that have misspelled words are also removed. Finally, 26 replies including ``\textit{supreme court regain conservative tilt with Gorsuch confirmation}'' remain.

\subsection{Neural Dialog Generation}
We used sequence-to-sequence variant approaches to generate candidate response. We tried different data sources to train different models: the Cornell movie lines corpus (\cite{cornell2011}), Reddit data, and Twitter data. We also coordinated our models with a topic detector, training different models using different topic corpora. We used character-based (\cite{sutskever2011generating}) and word-based sequence-to-sequence  (\cite{klein2017opennmt}) models. The word-based sequence-to-sequence architecture  consisted of 2 layers of Recurrent Neural Networks (RNN) with 500 hidden units. We used 2 layers of LSTM in our character-based models, with 128 internal states in the LSTM. The encoder of our model is just the RNN, while the decoder applies attention over the source sequence. We use a front-end to generate responses that invoked the different models.

\subsection{Engagement Reranking}
Sequence-to-sequence models tend to generate generic non-engaging responses such as ``\textit{I don’t know}'' or ``\textit{Thanks for your suggestions}'' (\cite{yu2015ticktock}). Collecting candidate responses from multiple neural dialog generation sources and reranking them is one solution. We create an engagement-based model to re-rank all candidate responses. For the training set, we separate Reddit commentaries into two groups: engaging (commentaries with scores larger than 200) and non-engaging (comments with score of 0). We then extract lexical features  (bag of words, commentary length, duplicate ratio) and external features (elapsed time between the commentary and its post, upvotes of the post, overlap ratio between the commentary and its post) to train a classifier. We compared three classification models: (1) logistic regression, (2) linear SVM classifier and (3) multinomial naive Bayes classifier. 
For both training set and testing set, we use 3000 positive and 3000 negative instances. These three models have similar performance, while the linear SVM classifier performs best, giving  79.7\% accuracy on the test set for binary classification,  comparable to previous work \cite{terentiev2014predicting}. We use the confidence score to rerank response candidates. 

This approach has some drawbacks. First, as we use the model for reranking instead of binary classification, the accuracy mentioned above may not reflect actual performance. Second, the external features contribute much more than the lexical features: alone they give about 75\% accuracy while using lexical features alone only leads to about 60\% accuracy.  This is understandable because the popularity of a commentary may be highly correlated to the popularity of its parent post, post time and initial interactions, but less correlated to its own content. Another possible reason is that bag-of-words model may be too naive to capture engagement level.  
To overcome these two issues, we plan to investigate different objective functions and evaluation metrics for  reranking. More complex features, such as bigrams and trigrams, or an LSTM model could also be used for reranking.

\subsection{Context Tracking}
Context tracking is used for coreference resolution. When a sentence from the user appears, we retrieve the most recent (with a default of five) chatting records of that user from the chat history database. The Stanford CoreNLP toolkit (\cite{manning2014stanford}) is used to resolve coreference. The pronouns in the new sentence are replaced if a coreferent is identified.

One difficulty is that there can be multiple references for one pronoun, only some of which are suitable as a replacement. For example, if a user first asks: ``\textit{Do you know France?}'', the socialbot replies: `\textit{`France, officially the French Republic, is a country with territory in western Europe and several overseas regions and territories.}''. The user may then ask: ``\textit{What’s the capital of it?}'' then. All of ``\textit{France}'', ``\textit{the France Republic}'' and ``\textit{a country}'' coreference with ``\textit{it}'', but it is not helpful to replace ``\textit{it}'' with ``\textit{a country}''. We currently use heuristic resolution and choose the mention that appears first in the context (i.e. \textit{``France'' here}): we observe that people usually use a clear and explicit mention when an item is firstly referred.

Poor coreference resolution can lead to problems. If a user says: ``\textit{I've never been France.}'', and the system replied ``\textit{thank you for being the most incredible and inspiring woman i\'ve ever met love from France I miss you, can't wait til June}''. The user may then ask ``\textit{what's the capital of it}''. The coreference resolution would mistakenly replace ``\textit{it}'' with ``\textit{June}''. A better model (say neural-based) might improve performance   co-reference resolution, but at the cost of more computation and consequent response longer delay.

\subsection{Miscellaneous Features}
Conversations may have certain predictable features that are better dealt by using specific strategies; accordingly, we introduced several ad hoc procedures to RubyStar.

\subsubsection{Back-story}
In order to avoid controversial conversation topics such as politics, pornography and etc, we gave RubyStar the personality of a nine-year-old girl and provided a series of back-story facts and opinions. 
We match a user input to the back-story database, if the score is above a threshold threshold, we enter the back-story context and reply using a suitable template. 

\begin{wrapfigure}{r}{0.3\linewidth}
\centering
\vspace{-10pt}
\includegraphics[scale=0.25]{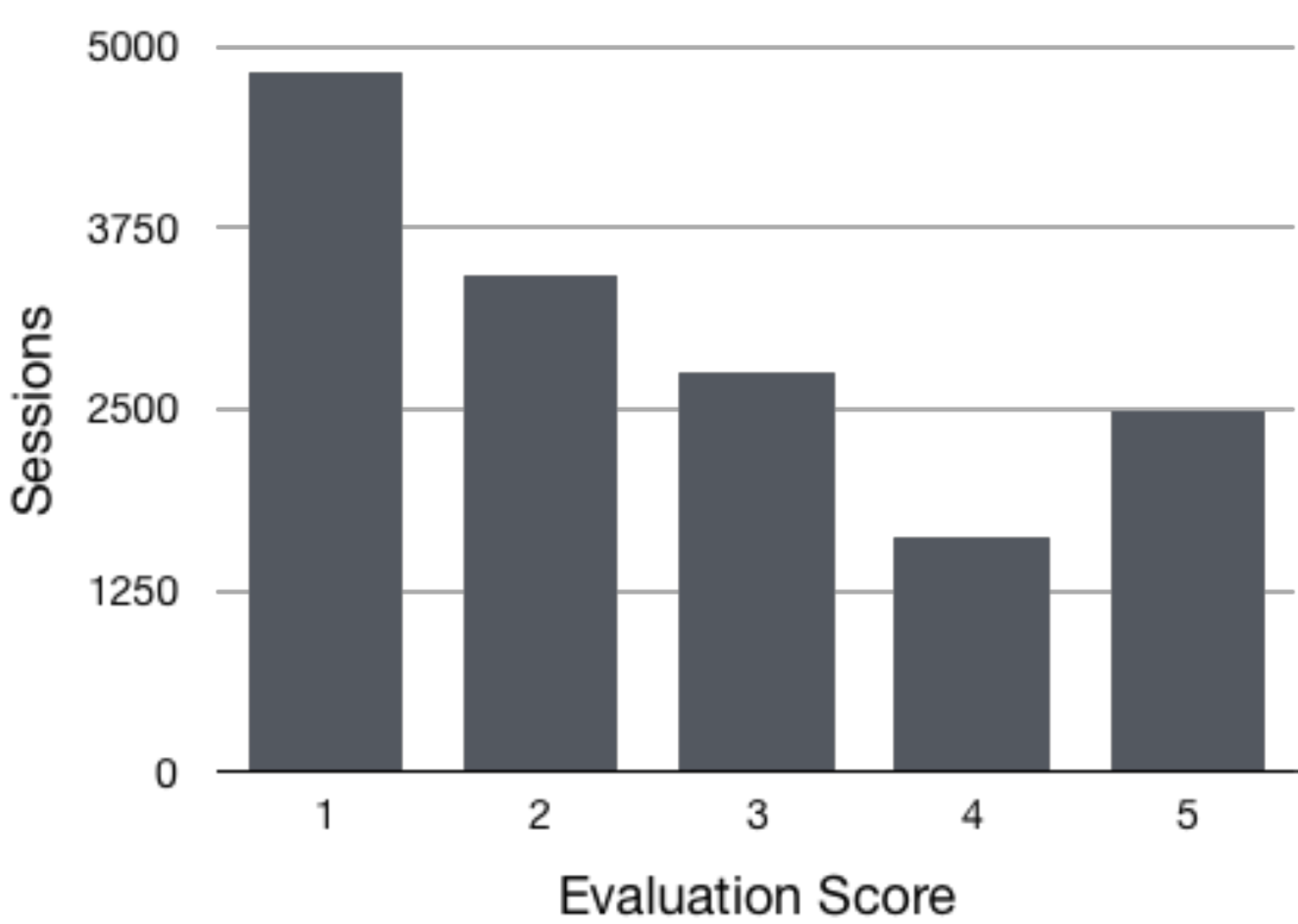}
\caption{Score distribution during the Alexa Contest} 
\label{figure2}
\vspace{-34pt}
\end{wrapfigure}

% \subsubsection{Profanity}
% Using publicly available data for response generation creates certain problems, for example abusive language. We've observed it in out Reddit and Twitter datasets.
% To assure acceptable behavior we used several criteria. First we filtered out responses using a banned-word list. Second we use a model-based approach to predict a sentence’s profanity level. The model consisted of an ensemble of many simple classifiers including SVM, naive Bayes, and Logistic Regression. We used this filter on the training data and also as part of the response candidate selection process.

\subsubsection{Question Answering}
Some user inputs will be factual questions. When a strong question asking pattern is detected in the input we pass it to Amazon Evi \footnote{\url{https://www.evi.com/}}.  
If a meaningful answer (rather than ``I don’t know'') is returned, we use it as our  response.

\subsubsection{Chat History and Customer Feedback}

The system is instrumented to collect performance data; we record intermediate process data and results to support debugging, analysis, and profiling. The log provides the time spent in each module, templates matched, candidate replies and so on.

History logging is a good resource to profile and identify the underlying problems in efficiency. We run the system on t2.medium machines on Amazon EC2. Table \ref{examples} and Table \ref{examples-bad} show some example conversations users may have if they chat with our system. We use profiling to identify significant problems; we would use this process to optimize response lag time. The actually delay of replies can be different from the sum of all of the modules, because (1) some module can be skipped, e.g. if it hits the backstory, the result will be returned without other steps; (2) some steps can be parallelized e.g. different neural network modules.

We log conversation transcripts with timestamps and additional meta data. This chat history database is queried and used for purposes of determining context. A web user interface is also available for reviewing conversations

% \subsubsection{Customer Feedback}
During the evaluation period, at the end of the conversation, the system asked users to give a rating for the conversation on a scale of 1 to 5. We leveraged customer feedback to improve our socialbot mainly in two ways. On the one hand, we use this as a source of inspiration for new strategies. On the other hand, we also use the ratings to test whether our new strategy improve the customer experience. By looking into the conversations with low ratings, we found that many could be handled by detecting common patterns. As a result, we introduced the intent templates module (section 3.1). We also found that many low rating conversations did not have smooth transitions. To address this, we added an open question at the end of each intent template. These two strategies improved the daily average rating by 10\% (on the 5-point scale).

\section{Analysis}
There is little current consensus on how to systematically evaluate the quality of conversational non-task interaction by socialbots. Although quantified evaluation methods like BLEU score (\cite{papineni2002bleu}) and METEOR (\cite{banerjee2005meteor}) are commonly used for machine translation, summarization and related fields, they are of limited use for evaluating chats (\cite{liu2016not}). We used human evaluation provided by the Alexa Prize evaluation to assess system performance.  The RubyStar developer team describes the quality of RubyStar responses using a 1 to 5 scale. The meanings of 1 to 5 are: Not human readable reply, Understandable and somehow related reply, Acceptable but not very meaningful and not very engaging reply, Good not very engaging reply, Excellent engaging and meaningful reply.

\begin{wrapfigure}{R}{0.4\linewidth}
\centering
\vspace{-22pt}
\includegraphics[scale=0.4]{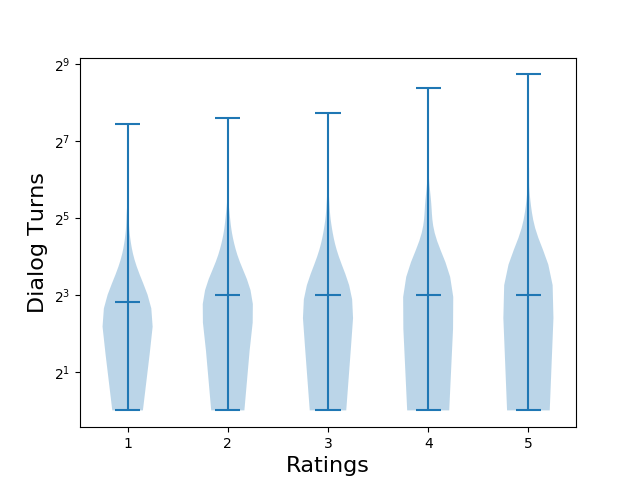}
\caption{The distribution of dialog turns for sessions with different rating via violin plot. The blue bars in the middle indict median values.} 
\label{violin}
\vspace{-20pt}
\end{wrapfigure}

We collected statistics for responses from Alexa users who interacted with RubyStar between May 15th 2017 and June 30th 2017; a total of 15084 sessions were collected. These sessions have an average score of 2.57; 45.5 \% of our test conversation is scored 3 or higher (Figure \ref{figure2}). We believe this demonstrates RubyStar’s basic capability of understanding non-task oriented utterance and giving appropriate reply, though various improvements seem advisable. For example have RubyStar take a more active role in managing the direction of the conversation, and better balance the effort needed to continue. 

Figure \ref{violin} shows the distribution of dialog turns under different ratings. We find high-rating sessions have longer tails in the distribution, but the median dialog turns are similar between different ratings. The result implies that dialog turns are not the dominant indicators for session ratings. The trend can also be viewed at the top of Table \ref{rating}.

Table \ref{examples} and Table \ref{examples-bad} show some examples generated by RubyStar. In satisfactory examples, RubyStar does entity recognition very well and replies with an engaging utterance. In unsatisfactory examples, reasons include (1) failure of reference recognition (2) failure to do context tracking between different generators (3) failure to understand the semantics of the conversation.

\begin{table}
\begin{center}
\caption{Examples of good exchanges{\label{examples}}}
% \medskip
\begin{tabular}{|p{0.95\columnwidth}|}
\hline 
\tiny {\textit{do you have emotions.}}
\newline 
\tiny \texttt{(evi) I suspect you know the answer to that already.}
\newline
\tiny {\textit{so you don't have emotions.}}
\newline 
\tiny \texttt{(twitter) 	you know that feeing where you wanna cry but you don't want your parents to notice so you just hold it in until they go to bed.}
\newline
\tiny {\textit{why do you not have emotions.}}
\newline 
\tiny \texttt{(neural) I don't know how to describe it. If you don't agree with me, I'll have a hard time believing you're not.}\\ \hline 

\tiny {\textit{is it possible that I can make you smarter}}
\newline
\tiny  \texttt{(neural) Of course! I am going to try to stay up too late today and tomorrow morning. Thanks for the help!}
\newline
\tiny {\textit{no problem I'm always here for you}}
\newline
\tiny  \texttt{(twitter)  may not be right there whenever you needed me , but just know that i'm always here for you}
\\ \hline 
\tiny {\textit{am I worthless}}
\newline
\tiny \texttt{(neural) I like the way you think. I didn't say that. I didn't mean to be rude. I just wanted to tell you that you don't need to have an actual point.}
\newline
\tiny {\textit{well I have an actual point my point is that love is all that matters.}}
\newline
\tiny {\texttt{(twitter) and when you figure out , love is all that matters after all , it sure makes everything else seem so small.}}
\\ \hline 
\end{tabular}
\end{center}
\end{table}

Table \ref{rating} shows the number of turns correlates with rating: higher ratings are associated with  longer conversations. A user enjoying the interaction will spend more time on it; the compressed median range implies that enjoyable conversations can last for some time. We also show some marker words which have correlations with ratings. We find the word "love" and "friend" is positively correlated with higher ratings. On the other hand, the word "stupid" has negative correlation with higher ratings.

\begin{table}
\begin{center}
\caption{Examples of unsatisfactory exchanges{\label{examples-bad}}}
\medskip
\begin{tabular}{|p{0.95\columnwidth}|}
\hline 
\tiny{\textit{do you like dogs}}
\newline 
\tiny{\texttt{(evi) I don't have a dog, but if I did, I would name him Astro!}}
\newline 
\tiny{\textit{cool I dog his name is Louis}}
\newline 
\tiny{\texttt{(twitter) i don't drink coffee or eat avocado and i dog even have enough to save for a dolls house}}
\newline
\tiny{\textit{  that's not what I asked}}
\newline 
\tiny{\texttt{(evi) Sorry, I know a lot but I don't know everything.}}
\newline
\tiny{\textit{I have I told her that I have a dog named Louis}}
\newline 
\tiny{\texttt{(neural) Do you have anything else?}}
\newline
\tiny{\textit{I have a dog}}
\newline 
\tiny{\texttt{(evi) I would guess your dog's name is Fido! Give them a pet for me, even if it's not.}}
\\ \hline
\tiny {\textit{what do you think a wonder woman}}
\newline 
\tiny \texttt{(twitter) alright i just watched wonder woman and i cried.}
\newline
\tiny {\textit{what was your favorite thing about wonder woman}}
\newline 
\tiny \texttt{(twitter) Aha! shout out to chris for taking me to see wonder woman.}
\newline
\tiny {\textit{chris who is chris}}
\newline 
\tiny \texttt{(evi) Chris is a short form of various names including Christopher, Christian, Christina, Christine, and Christos.}\\\hline 
\end{tabular}
\end{center}
\vspace{-15pt}
\end{table}

Table \ref{rating} also shows the usage ratio of different response generators. We can see that Neural Models accounts for more than 40 percent generated responses. Notably, high-rating dialogues have fewer backstory-related generations, but more Twitter generations. This also suggests a possible way to produce more coherent dialogues.

\begin{table}[h]
\begin{center}
\caption{\label{rating} Statistics for dialogs across different user rating levels}
\medskip
\small
\begin{tabular}{ l*{5}{S[table-format=4.2, table-number-alignment=center]}} %{|l|lllll|}
\hline \bf Statistics & \bf Rating:1 & \bf Rating:2 & \bf Rating:3 & \bf Rating:4 & \bf Rating:5 \\ \hline
Mean Dialog Turns & 10.0 & 12.9 & 13.4 & 16.0 & 16.6 \\
Median Dialog Turns & 7 & 8 & 8 & 8 & 8 \\
\hline
Sessions with  "love" & 13.4 \% & 17.4 \% & 20.9 \% & 21.0 \% & 23.3 \% \\
Sessions with  "friend" & 14.2 \% & 17.6 \% & 19.5 \% & 24.4 \% & 20.7 \% \\
Sessions with  "hate" & 5.3 \% & 6.3 \% & 6.2 \% & 8.9 \% & 8.3 \% \\

\hline \bf Generator & \bf Rating:1 & \bf Rating:2 & \bf Rating:3 & \bf Rating:4 & \bf Rating:5 \\ \hline

Backstory & 9.4\% & 9.3 \% & 9.7 \% & 8.3 \% & 7.3 \% \\
Neural & 44.5 \% & 42.6 \% & 40.6 \% & 39.2 \% & 41.0 \% \\
Evi & 23.1 \% & 24.4 \% & 26.4 \% & 28.4 \% & 26.9 \% \\
Twitter & 20.7 \% & 20.7 \% & 20.4 \% & 21.6 \% & 22.3 \% \\
Rule & 2.4 \% & 3.0 \% & 2.9 \% & 2.4 \% & 2.5 \% \\

\hline
\end{tabular}
\end{center}
\vspace{-10pt}
\end{table}

We received more than 1300 feedbacks from Alexa users. Positive comments mention ``humorous'' and ``funny''. Complaints fall into two main clusters. The first is about content: \textit{``Her responses were  reactive and not prompting anything which made it very difficult to have a conversation''} or \textit{``At one point it told me it didn't care about my life. And I was felt very negative''}.  The personality was described as ``weird'' or ``creepy''. The other is about coherence: \textit{``[RubyStar] Suggested a topic but couldn't go any further. Really frustrating''} and \textit{``It asked about food and then didn't know what i was talking about when i talked about food''}.  RubyStar doesn't provide explicit guidance; this is confusing:  \textit{``It doesn't converse it just has its own agenda and I have no idea how to converse with it''}.

\section{Conclusion and Future Work}
We described the architecture of the RubyStar socialbot. We found that difference response generators has pros and cons, based on different conversation scenarios. By combining rule-based, retrieval-based and generative approaches and enhancing them with the use of context tracking as well as engagement reranking, our system performs better in the experimental evaluation. 

RubyStar currently provides only limited context tracking and conversation level coherence. One remedy might be to add a GANs ~(\cite{li2017adversarial}) model to do reply generation and use reinforcement learning methods~(\cite{li2016deep}) to select a reply based on the current state of the conversation. Word embeddings could be used to replace the current one-hot encoding and augmenting it with topic embedding, sentiment embedding and engagement embedding. Engagement management and sentiment analysis could improve conversational strategies. For the Information Retrieval module, we would investigate training a language model based on conversation transcripts, and using perplexity to evaluate tweets instead of simply using misspelled word counts. we believe that a learning-based approach to ranking will help select the most suitable tweets for a reply. Finally, actively managed participation would make the system easier to engage with.

% RubyStar, as well as other socialbot systems are being made publicly available as part of the Alexa Prize competition\footnote{\url{https://developer.amazon.com/alexaprize}}. 

% \clearpage

\bibliography{ap_2017}
\end{document}